\documentclass[sigconf]{acmart}

\usepackage{booktabs} 

\setcopyright{rightsretained}
\usepackage{graphics} 
\usepackage[caption=false,font=footnotesize]{subfig}

\allowdisplaybreaks

\title[Distribution Invariant Regression Metrics]{
A First Step Towards Distribution Invariant Regression Metrics
}

\begin{document}

\author{Mario Michael Krell}
\authornote{This work was supported by a fellowship within the FITweltweit program
of the German Academic Exchange Service (DAAD)}
\affiliation{%
 \institution{International Computer Science Institute, Berkeley}
 \streetaddress{1947 Center Street}
 \city{Berkeley}
 \state{California}
 \country{USA}}
\affiliation{%
  \institution{EECS Dept., University of California, Berkeley}
  \city{Berkeley}
  \state{California}
  \country{USA}
}
\affiliation{%
  \institution{Robotics Research Group, University of Bremen}
  \city{Bremen}
  \state{Bremen}
  \country{Germany}
}

\author{Bilal Wehbe}
\authornote{This work was supported by the Marie Curie ITN program ``Robocademy'' FP7-PEOPLE-2013-ITN- 608096.}
\affiliation{%
  \institution{Robotics Research Group, University of Bremen}
  \city{Bremen}
  \state{Bremen}
  \country{Germany}
}
\affiliation{%
  \institution{Robotics Innovation Center, DFKI GmbH}
  \city{Bremen}
  \state{Bremen}
  \country{Germany}
}

\begin{abstract}

  Regression evaluation has been performed for decades.
  Some metrics have been identified to be robust against shifting
  and scaling of the data but
  considering the different distributions of data 
  is much more difficult to address (imbalance problem)
  even though it largely impacts the comparability between evaluations
  on different datasets.
  In classification, it has been stated repeatedly that 
  performance metrics like the F-Measure and Accuracy are highly dependent on
  the class distribution and that comparisons between different datasets
  with different distributions are impossible.
  We show that the same problem exists in regression.
  The distribution of odometry parameters in robotic applications
  can for example largely
  vary between different recording sessions.
  Here, we need
  regression algorithms that either perform 
  equally well for all function values,
  or that focus on certain boundary regions like high speed.
  This has to be reflected in the evaluation metric.
  We propose the modification of established regression metrics
  by weighting with the inverse distribution of function values $Y$
  or the samples $X$ using an automatically tuned Gaussian
  kernel density estimator. 
  We show on synthetic and robotic data in reproducible experiments
  that classical metrics behave wrongly, whereas our new metrics
  are less sensitive to changing distributions,
  especially when correcting by the marginal distribution in $X$.
  Our new evaluation concept enables the comparison of results
  between different datasets with different distributions.
  Furthermore, it can reveal
  overfitting of a regression algorithm to overrepresented target
  values.
  As an outcome, non-overfitting regression algorithms 
  will be more likely chosen
  due to our corrected metrics.
\end{abstract}

%


\maketitle

\section{INTRODUCTION}

\begin{figure*}[thpb]
  \centering
  \includegraphics[height=0.2\linewidth]{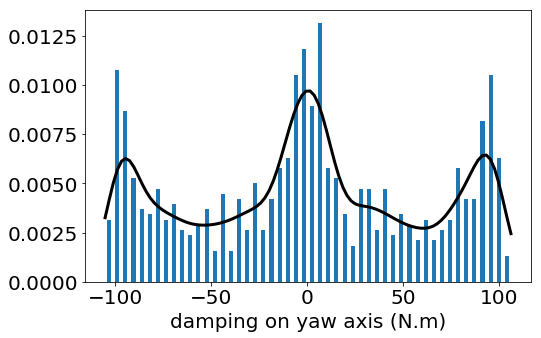}
  \includegraphics[height=0.2\linewidth]{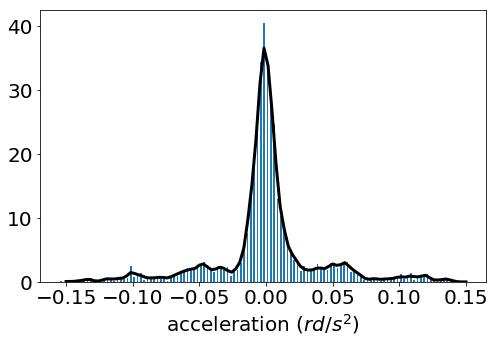}
  \includegraphics[height=0.2\linewidth]{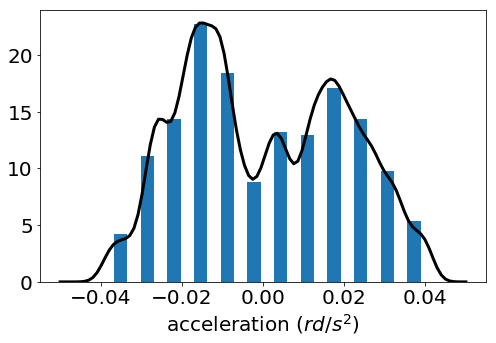}
  \caption{Histograms from robotics data (unmanned underwater vehicle)
  showing non-uniform distributions of target values
  and respective fitted kernel density estimations.} 
  \label{f:KDE}
\end{figure*}

Knowledge discovery for robot is becoming increasingly complex and learning algorithms
like regression have to be used to model their dynamics or environment.
Consider for example a case of an autonomous vehicle where we record data and
want to estimate certain parameters 
of its odometry model like velocity and acceleration.
Due to external factors,
we might have more, small accelerations in one experiment and
more, high accelerations in
another experiment 
(see also Fig.~\ref{f:KDE} for examples with non-uniform distributions).
We are interested in regression algorithms
that are either equally good for all accelerations 
or prioritize higher accelerations.
However, an evaluation on the dataset with small accelerations 
could produce excellent mean-squared error (MSE),
even though the algorithm might be completely off for higher accelerations.
The reason could be that high accelerations did not occur that often
in the dataset and that the dynamics in those regions are more complicated.
When the vehicle moves slowly, parameter relationships might 
behave linearly whereas for higher accelerations, 
nonlinear relationships have to be considered. 
So the evaluation metric 
might prefer a linear algorithm in contrast to a nonlinear one
because it matches the overrepresented small accelerations slightly better.
But the nonlinear approach might work perfectly for high accelerations
where the linear approach fails.
Apart from this, there is the issue 
that evaluation results on a different dataset will not
be comparable because the underlying distributions of 
samples $X$ and function values $Y$ 
are completely
different (also called imbalance problem).
We claim that good performance metrics should be independent from
dataset properties to enable a good comparison of the regression algorithm.
We suggest to compensate for those distribution differences by using
a \emph{correction by inverse weights from kernel-density estimates}.

This approach is closely related 
to importance sampling in statistics~\cite{Neal2001}.
Given data with a density $q(x)$, the expected value of a function
$f(x)$ is estimated by sampling $x_1,\ldots,x_N$ with $q$ 
and then calculating:
$E[f|q]=\frac{1}{N}\sum f(x_i)$.
The simplified basic idea of importance sampling is 
to sample with a density $g(x)$ instead and correct for it
by the weighted average
\begin{align}
E[f|q] = \frac{\sum f(x_i)w_i}{\sum w_i} \text{ with weights } w_i=\frac{q(x_i)}{g(x_i)}\, .
\end{align}
Importance sampling is usually applied in Monte Carlo methods
to calculate an expected value more efficiently or reduce the variance.
The main difference to our application is that we don't look
only at expected values, but rather at different error functions.
Furthermore, the evaluation dataset has already been chosen by the
data recording. 
Hence, it is not possible to modify it to calculate the error function
more efficiently or reduce the variance---usually all data is used 
for evaluation.
However, from the perspective of importance sampling,
metrics like MSE are calculated in the wrong way, assuming
data from $X$ has been sampled with density $g$ 
which is different from the desired density $q$. 
The density $q$ could represent the ``true density'', 
the density of a different dataset that we want to compare with,
or a density that describes the relevance for the application.
Practically, we don't want to sample from a different density function
but use the estimated density of the data being given to correct for
the bias in the regression metrics.

\textbf{Definition: Distribution Invariant Metric}
Let $D_1$ and $D_2$ be two differently distributed datasets
that cover the whole range of interest for the application, 
$A_1$ and $A_2$ be the true function values,
$P_1$ and $P_2$ be the predictions  coming from the same regression function
(trained on some other data beforehand),
and $f$ be a regression metric. 
We say $f$ is \textbf{distribution invariant},
if and only if $f(D_1,A_1,P_1)=f(D_2,A_2,P_2)$
for any choice of $D_1$ and $D_2$.

We are assuming that the true underlying functional relationship
stays the same, namely the probability of $P_i$ given $D_i$.
This makes this topic slightly different to
classical covariate shifts and transfer learning~\cite{Candela2009}.



An overview over the imbalance problem in classification, 
as well as regression,
is provided by Branco et al.~\cite{Branco2016}.
They focus on the application perspective,
where some prediction values have more relevance/costs than others.
A major part of the literature introduces 
asymmetric loss functions to differentiate between positive and negative errors
like a ROC curve for regression~\cite{Hernandez-Orallo2013}
and the LIN-LIN loss~\cite{Christoffersen1996,Crone2005}.
For balancing the data, the SMOTE approach for regression 
has been introduced~\cite{Torgo2013}.
For transferring a regression function between different contexts,
a reframing concept was presented~\cite{Hernandez-Orallo2014}
where the trained model was adapted to new loss functions.
This approach does not handle different distributions in the data.
A highly related topic is the area of cost sensitive learning.
A straightforward approach is to introduce weights in the classical metrics
to emphasize certain errors.
Still, it remains open how to define those
weights and how to handle different $Y$-distributions.
The same holds for the approaches by 
Ribeiro et al.~\cite{ribeiro2011utility,Torgo2009}.
They point out that for some applications, rare values in $Y$ might be very
important and there should be very low error in their prediction.
Furthermore, false predictions
that assign those rare values to samples where the true value is not rare,
should be avoided.
To solve this problem, the user has to define a special handcrafted relevance
function and a threshold for calculating metrics that were inspired
by precision and recall.

In contrast, we propose a parameter free solution 
that also generalizes over evaluations 
on different datasets and makes them better comparable.
So far, the imbalance problem has not been tackled from the 
distribution perspective by reusing existing metrics.
Especially, the lack of comparability between evaluations with different
distributions has not yet been addressed at all.
Note, different $Y$-distributions are a result of the 
mapping function as well as the underlying distribution of samples ($X$).
Hence, we compare both approaches.
The $X$-perspective is motivated by importance sampling and the
$Y$-perspective by classical approaches for metrics.
Apart from evaluation metrics, our approach can be also applied
to loss functions as they occur in most regression algorithms.

In Section~\ref{s:sota}, we review the related work.
Our method is introduced in Section~\ref{s:meth}.
For the evaluation in Section~\ref{s:eval}, 
we first explore the required kernel-density estimation
(KDE, Section~\ref{s:KDEE}) and then we look at specifically designed
synthetic data examples (Section~\ref{s:synth}), as well as
a real world application on robotic data (Section~\ref{s:rob}).
Finally, we provide a conclusion and discuss future steps
in Section~\ref{s:conc}.

\section{Related Work}
\label{s:sota}

Looking into the literature, it turns out that classification problems
are much more prominent in the machine learning literature
then regression problems.
For classification tasks, the problem of data 
imbalance/bias~\cite{Branco2016,Kordopatis-Zilos2016,Candela2009,Sun2009,Torralba2011} and
its effect on the metrics has been discussed exhaustively
(see Straube et al. \cite{Straube2014} for a review).
Kubat et al. \cite{Kubat1998} point out that accuracy is not a good 
evaluation metric
for data with imbalanced class ratio because good results can be obtained 
with a constant prediction of the overrepresented class 
(areas without oil spills in their case).
Straube et al. emphasized that the resulting famous F-Measure is not
a good way out because it is highly sensitive to the class ratio.
This was confirmed 
from a different perspective by Lipton et al.~\cite{Lipton2014}.
Possible solution approaches in classification 
are metrics that are based on rates like
geometric mean, balanced accuracy, or area under roc curve
\cite{Bradley1997,Kubat1998,Straube2014}.
These methods treat the two different classes equally.
An approach from the data perspective is to oversample the underrepresented
class or reduce the overrepresented class, which has a similar effect.

So far, there is few literature 
that tackles the 
much more challenging counterpart in the regression area.
For regression, we do not have two classes, but we
have a distribution of the dependent variable, $Y$, as well as
the independent variables, $X$.
The task of the regression model is to construct a function $f$
that approximates $f(x_i) = y_i$. 
Depending on the application, the distribution of $X$ and $Y$ cannot
be assumed to be fixed or uniform.
Note, that a uniform distribution would be the generalization
of the balanced class ratio in a classification problem.


\subsection{Regression Metrics}
\label{s:metrics}

Relevant regression metrics have been summarized 
by Witten et al. \cite{witten2016data}.
The most widely used metric is probably the means-squared error (MSE).
This is the error minimized in linear regression,
which assumes that the residuals are Gaussian distributed.
It is also commonly used for neural networks.
Given the predictions $p_i$ and the actual measured values $a_i$ from
a training set $x_i$ with $n$ samples it is defined as
$   \mathit{MSE}(a,p) = \frac{1}{n}\sum 
    (p_i-a_i)^2
    \text{, where } \mathit{RMSE}(a,p) = \sqrt{\mathit{MSE}(a,p)}$
is the root MSE (RMSE).
Instead of squares it is also possible to use absolute values
for the mean absolute error (MAE)~\cite{Willmott2005}:
 $   \mathit{MAE}(a,p) = \frac{1}{n}\sum 
     \left|p_i-a_i\right|$.
For scaling invariance and better comparability, 
several other metrics have been suggested which use the average of 
the actual values 
$\left(\bar{a}=\frac{1}{n}\sum a_i\right)$\cite{witten2016data} like
the relative squared error (RSE),
the root RSE ($\mathit{RRSE}(a,p) = \sqrt{\mathit{RSE}(a,p)}$), 
and the relative absolute error (RAE):
\begin{align}
    \mathit{RSE}(a,p) &= \frac{\sum\limits_{i=1}^n (p_i-a_i)^2}{
            \sum\limits_{i=1}^n (a_i-\bar{a})^2},\,
    \mathit{RAE}(a,p) = \frac{\sum\limits_{i=1}^n \left|p_i-a_i\right|}{
            \sum\limits_{i=1}^n \left|a_i-\bar{a}\right|}\, .
\end{align}
It is common for classification, 
to calculate metrics that correspond to random guessing 
or estimating just one class.
The respective counterpart in regression is to look at the average
actual values and consider $p_i\equiv \bar{a}$ as the worst case.
Whereas for linear regression, the relative metrics (RSE, RRSE, RAE)
should not exceed $1$ due to this worst case, this can happen for nonlinear
regression problems.

Apart from those straightforward sums, there are also metrics that come
from statistics.
The (Pearson) correlation coefficient (PCC)~\cite{pearson1895note}, takes the correlation of 
actual and predicted values (relative to the mean values) and
normalizes the results by the respective autocorrelations:
\begin{equation}
    \mathit{PCC}(a,p) = \frac{\sum\limits_{i=1}^n (p_i-\bar{p})(a_i-\bar{a})
    }{\sqrt{\sum\limits_{i=1}^n (p_i-\bar{p})^2\cdot
            \sum\limits_{i=1}^n (a_i-\bar{a})^2}}\, .
\end{equation}
If the values are anti-correlated, values of $-1$ are possible,
whereas the maximum value is again $+1$.

Another metric from statistics is the $R^2$ score or 
``coefficient of determination'' (COD)~\cite{Carpenter1960},
which is usually defined as
$\mathit{COD} = 1-\mathit{RSE}$.
COD is a statistic for linear regression but widely used for any
regression evaluation, 
whereas the generalization to nonlinear regression~\cite{ColinCameron1997}
is rarely used.
If the prediction is worse than the average, COD can obtain large negative
numbers lower than $-1$ in the nonlinear case. 
Otherwise, it provides scores 
between $-1$ for the worst result and a maximum of $+1$ for the best result.
If the prediction is always constant disregarding the input features, COD
would then give a score of $0$.
The explained variance score (EVS) is similar but corrects the nominator of
COD by the mean, as it is common for the variance~\cite{Kent1983}:
\begin{equation}
    \mathit{EVS}(a,p) =1- \frac{
            \sum\limits_{i=1}^n ((p_i-a_i)-(\bar{p}-\bar{a}))^2
             }{ 
            \sum\limits_{i=1}^n (a_i-\bar{a})^2}\, .
\end{equation}

In this paper, we will not discuss the choice of an appropriate
metric but provide an approach to improve all of them using KDE.

%

\subsection{Kernel Density Estimation (KDE)}
\label{s:KDE}

Even though histograms are still quite famous,
they are usually not a good approach
to represent distributions because 
histograms can look completely different
depending on the underlying discretization (number of bins)~\cite{Pedregosa2011}. 
Instead, the more general KDE should be applied
to estimate the probability density function~\cite{Parzen1962,Rosenblatt1956}.
It can be considered as a smooth version of the histogram.
The general notation for a KDE is:
$g(a) = \sum\limits_{i=1}^n\frac{1}{nh}K\left(\frac{a-a_i}{h}\right)$
where the $a_i$ are usually assumed to be independent identically distributed.
The kernel function $K$ is non-negative, with zero mean and an integral of one.
Established examples are Gaussian, Tophat, Epanechnikov,
exponential, linear, and cosine kernel functions~\cite{Pedregosa2011}. 
The parameter $h$ defines the bandwidth and it is crucial to choose it
carefully.
There are several approaches implemented in Python libraries to 
automatically estimate this bandwidth.
Scipy~\cite{scipy} implements two simple heuristics called
Scott's rule~\cite{scott2015multivariate} $\left(n^{-1/(d+4)}\right)$ 
and Silverman's 
rule~\cite{silverman1986density}$\left(n \cdot (d + 2) / 4)^{-1 / (d + 4)}\right)$ 
where $d$ is the dimension of $Y$.
Note that for a good fit, it is required to adapt the bandwidth to the
data and optimize it.
Hence, 
cross-validation based optimization is usually a better choice than heuristics.
The statsmodels~\cite{seabold2010statsmodels} library provides
cross-validation,
using the maximum likelihood and the integrated mean square
error~\cite{li2007nonparametric}.

\section{Methods}
\label{s:meth}

For classification, a standard approach for imbalanced classification
is to determine the correct classifications for each class separately
and normalize them by the overall number of instances in this class
to obtain ratios like true positive/negative rate
and then average them in some sense (geometric/arithmetic mean).

A direct transfer to regression would be to 
discretize the $Y$ distribution into bins and
calculating the respective histogram.
Then, the regression results are divided by the number of occurrences 
in the respective bin.
This means that each bin gets the same relevance in the evaluation.
If the large majority of the data is in one bin where the regression
algorithm performs perfect but it is ``far'' off for the other bins,
this can be reflected by this measure whereas the common other measures
would basically ignore the other bins, if there is just enough data
in the correctly estimated bin.

Motivated by the state of the art in KDE
(Section~\ref{s:KDE}) and our evaluation in Section~\ref{s:KDEE},
we suggest to use this generalization (i.e., smooth version) 
of the histogram
with a Gaussian kernel and automatic bandwidth tuning
with efficient cross-validation using 
the maximum likelihood and the integrated mean square.
We chose the Gaussian kernel
because of its smoothing properties and general validity.
The KDE model is trained on the actual values $a_i$ in the
respective testing set.
Eventually, the results are weighted by the inverse of the
probability density function applied again to the respective $a_i$.
The predictions ($p_i$) should not be used 
because they should not influence
this part of the normalization.

Note that this maps the distribution of $a_i$ basically to a
uniform distribution.
If a uniform distribution is considered inappropriate, 
the weights can be corrected by multiplying it with the target
distribution $q$ afterwards as it is also done in importance sampling.
This way, too high weights on noisy values 
at the boundary region can be avoided.
The target distribution could be for example 
derived from the training set.
Another example in location estimation,
when you want to limit the evaluation to images from Paris.
Then you put zero weight to images from the USA
to compare with results on a different
dataset that is solely based on images from Paris.


Let $g$ be the density function, constructed from the $a_i$.
Our approach is now to reformulate all sums occurring in the 
definitions of metrics in Section~\ref{s:metrics}
to arithmetic means $\left(1/n\sum l(a_i, p_i)\right)$
and then
replace those by weighted means 
$\left(\sum g(a_i)^{-1}\cdot l(a_i, p_i)/\sum g(a_i)^{-1}\right)$
with the inverse distribution function values.
With $l(a_i,p_i)$ we denote any kind of occurring loss functions.
In some examples, there is a loss function in the nominator
as well as the denominatore, such that multiplying the respective
fraction with $\frac{1/n}{1/n}$ provides the given structure.
Hence, our new weighted COD would for example read:
\begin{equation}
    \mathit{COD}_{\mathit{g}}(a,p) =1- \frac{
        \sum\limits_{i=1}^n g(a_i)^{-1}\cdot(p_i-a_i)^2}{
        \sum\limits_{i=1}^n g(a_i)^{-1}\cdot(a_i-\bar{a})^2}\, .
\end{equation}

All other respective formulas can be found in the Appendix.
Similarly, a correction by the sample distribution ($X$) instead of using the
function values $a_i$ can be performed.
Therefore, the KDE $g$ is trained on the samples $x_i$ that correspond to
the true function values $a_i$ and $g(x_i)^{-1}$ is used instead of 
$g(a_i)^{-1}$.

Last but not least, this weighting can be combined with other weightings.
Approaches that address costs, priorities, or confidence could be incorporated
on top by weight multiplication. 
Furthermore, the distributions from other datasets could be used to make 
results comparable to the literature.

%

\section{Evaluation}
\label{s:eval}

First, we evaluate different KDE parameters to select an approach
for the weight correction.
Then we carefully analyze the properties of our new type of metrics on
synthetic data and eventually on real world data
for modeling the dynamics of an unmanned underwater vehicle (UUV).

All code is provided as
Jupyter notebooks in Python and data is uploaded
on Github~\cite{Supplement}.
The provided implementation of our new approach
uses existing libraries like statsmodels and NumPy
and can be directly integrated into
existing evaluation frameworks like scikit-learn~\cite{Pedregosa2011}
and pySPACE~\cite{Krell2013}.

\subsection{Analysis of Automatically Tuned Kernel Density Estimation}
\label{s:KDEE}
A general important advice, independent of the evaluation strategy,
is to look at the $Y$-distribution of the data as a histogram
as well as using a KDE.
This can also provide important insights into the data
and raise awareness of the differences between the $Y$-distribution
and a uniform distribution.

Since we prefer a smooth and non-vanishing density function,
we chose a Gaussian kernel, which showed sufficiently good approximation
capabilities on predefined distributions 
even with small sample size~\cite{Supplement}.
If the aforementioned visualization strongly suggests a different kernel,
we suggest to provide our weighted metrics 
normalized with the Gaussian kernel as well as
with the differently chosen one, to remain comparable.

In a preliminary analysis,
we analyzed the approximation capabilities and processing time
of KDEs on different distributions:
Gaussians, sum of Gaussians, Laplace, Chi square, and sum of uniform
and Gaussian distribution (see also Supplement~\cite{Supplement})
with the simple heuristics (Scott's and Silverman's rule)
and the cross-validation approaches.
In some cases, the simple heuristics performed much worse than the 
cross-validation approaches. Hence they are not a good choice,
even though they were much more faster with
around $0.02$ seconds, whereas the other approaches needed around $1$ second
for $1000$ samples with linear increase,
for simple distributions with only one peak. 
We also tested the efficiency parameter in statsmodels,
which performs the cross-validation on separate sub-samples of the data
to determine the bandwidth $h$.
For simple distributions, it provided no speed up
but turned out to be a good regularizer
for the cross-validation methods which sometimes showed random drops
in performance, when the efficiency option was not used.
For more complex distributions, like the sum of $3$ normal distributions,
the efficiency parameter is crucial 
($10$ instead of $75$ seconds for $10000$ samples).
If the time for the KDE fitting is crucial and too long due to
a large number of more than $10000$ samples,
Scott's rule could be used instead.

For the following evaluation, we focus on cross-validation with
the mean square error as metric instead of maximum likelihood,
because it slightly performed better in some cases.
For training the KDE, having only $100$ samples was sometimes 
too few and resulted in a performance below $0.9$ COD.
Whereas for $200$ samples, we always got a COD better than $0.9$
with our chosen method
with a slight increase when using more samples (up to $0.99$).
Depending on the data at hand, this approach has to be adapted,
when the KDE is too far off.

In Fig.~\ref{f:KDE}, we show the fitting of the chosen KDE approach
to target variable's (acceleration, damping) distribution 
(displayed as histogram) 
for three different robotic datasets from UUVs.

If the number of samples is large enough (at least$>1000$) 
and only a one-dimensional distribution
has to be modeled, it is also possible to use the histogram as a
density estimator instead.
Similar to the kernel width, there are heuristics that can determine the
optimal number of bins.
A brief comparison showed that from all $6$ heuristics in
numpy 
only the approach by Freedman et al.~\cite{Freedman1981} was sufficient
for noisy non-Gaussian distributions and was very close to the
estimated KDE and sometimes even better.

\begin{figure*}[t!]
\centering
\mbox{
  \subfloat[weighting by $Y$-distribution]{\includegraphics[width=\linewidth]{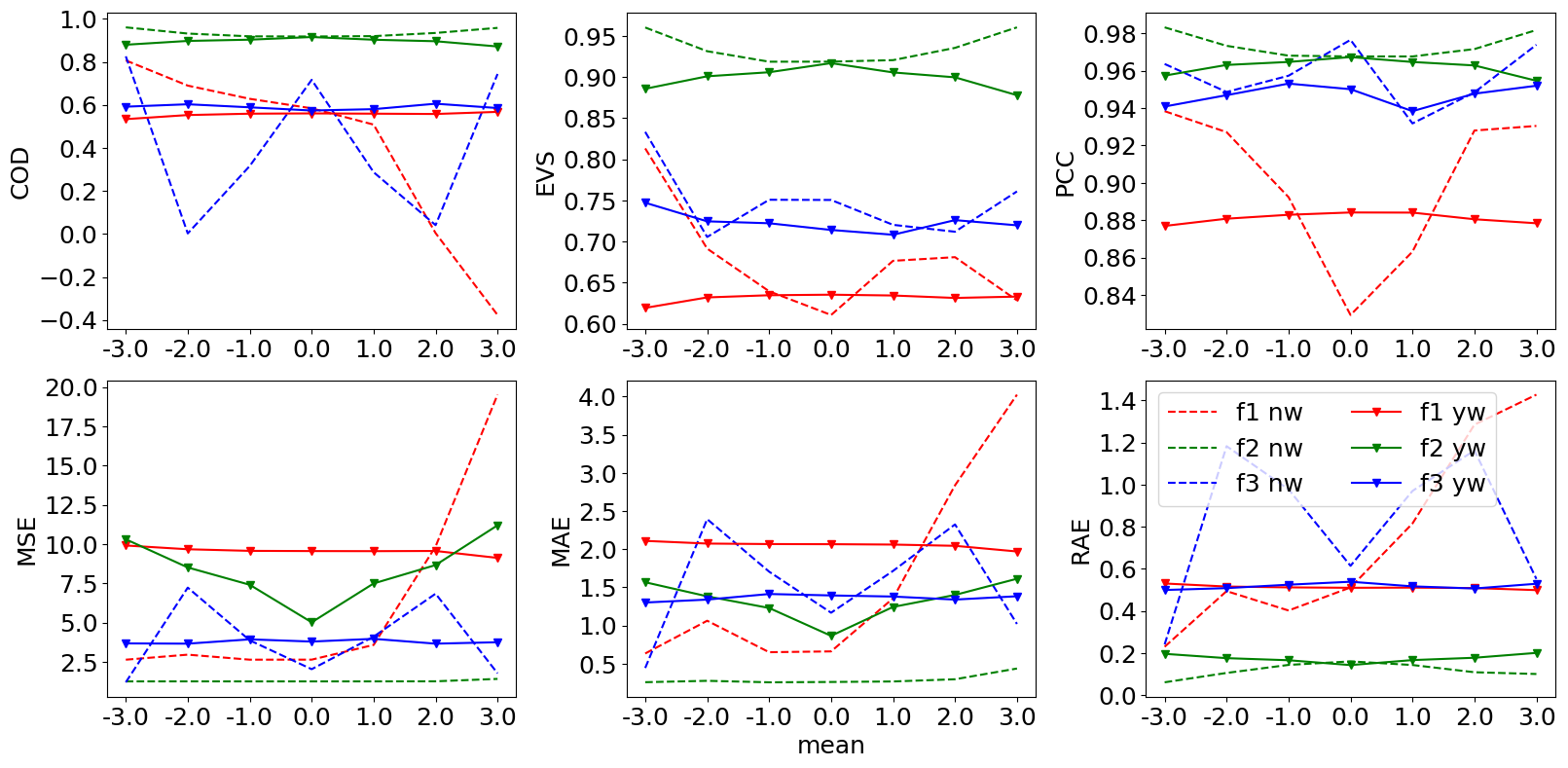}}
\vspace*{-2cm}
}

\mbox{
  \subfloat[weighting by $X$-distribution]{\includegraphics[width=\linewidth]{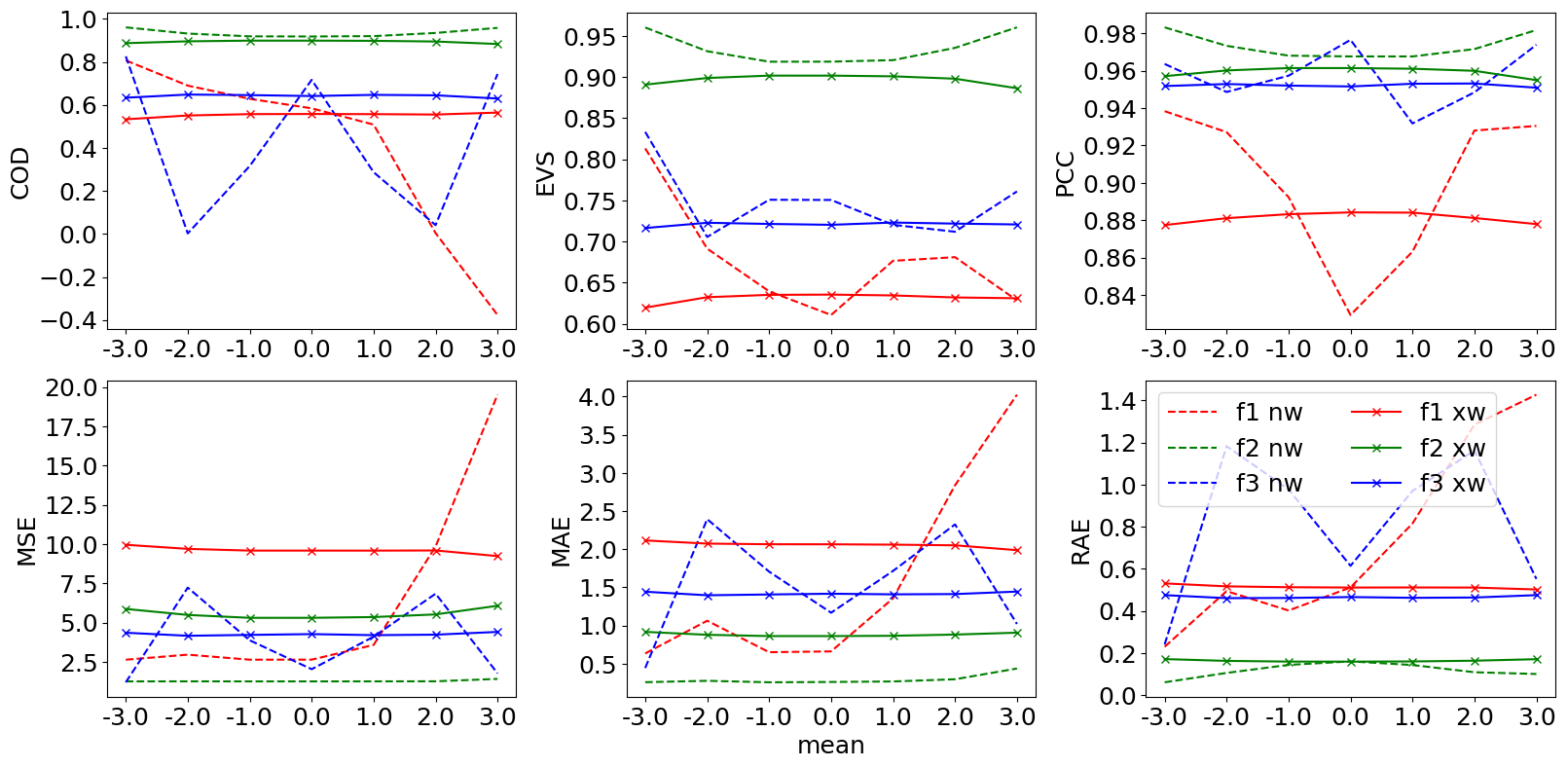}}
}
\caption{Comparison between classical metrics without weighting 
  (denoted by nw)
  and \emph{our} weight correction (denoted by yw and xw) 
  for three different synthetic functions (f1, f2, f3)
  using the Y-distribution (a) or the X-distribution (b).
  The x-axis displays the mean of the 
  moving normal distribution~(see Section~\ref{s:synth}).
  A distribution invariant metric would be constant for all means.}
\label{f:xyweighting}
\end{figure*} 

\subsection{Evaluation on Synthetic Data}
\label{s:synth}

For this analysis, we changed the distribution of $Y$ by changing the
distribution in $X$ and applying three functions of increasing
complexity
($2\cdot x$, $x\cdot|x|$, $10 \cdot cos^2(x)$) with random noise,
uniformly sampled from $[0,0.1]$.
The scaling factors ($2$, $10$) are used 
to map the MSE and MAE to similar ranges at the end.
For training we used, $300$ samples from a uniform distribution
from the interval $[-4,4)$ and $700$ samples from a normal distribution
with mean $-3$ and standard deviation $0.1$.
For comparing different distributions and the effect on established metrics
as well as our new modification, we consecutively shifted the mean by
$1$ up to a value of $3$ to obtain $7$ different testing sets~\cite{Supplement}.

As regression algorithm, 
we used support vector regression~\cite{Smola2004}
from scikit-learn~\cite{Pedregosa2011}
with $\epsilon=0.1$, kernel='rbf',  $\gamma=10$, and $C=0.1$.
For Function~2 ($x\cdot|x|$), we chose $C=0.5$, instead,
to get errors in a similar
range to simplify joint visualization with the other two functions.

The results are depicted in Fig.~\ref{f:xyweighting}(a).
They show the change of $6$ metrics depending on the shifted mean.
It can be seen that all classical metrics (without weighting correction)
change with the shifting mean in the testing sample distribution
but with different patterns for the functions as well as the metrics.
This proves the need for distribution invariant regression metrics.
Interestingly, our new correction of the metrics by the $Y$-distribution
worked very well for Function~1 and 3 but not that well for
Function~2 for MSE and MAE, 
for a mean value of $0$ for the $X$-distribution.
A possible reason for that might be that even for a uniform distribution of $X$,
we would get an accumulation of weights around $0$ in $Y$.
This indicates that the distribution of $X$ might be more relevant.

Hence, we also tested a correction in the weighting by the probabilities
of the underlying samples in $X$.
The results are depicted in Fig.~\ref{f:xyweighting}(b).
Now, the modified metrics are almost insensitive to the changing distribution.
Slight deviations might be due to inaccuracies in the KDE calculation,
because the chosen Gaussian kernel is not that well suited to
fit uniform distributions.
But note that in normal experiments, uniform sampling rarely occurs.

All in all, for one-dimensional function approximation, 
weighting based on the $X$-distribution
is the best approach to obtain distribution invariant performance values.
But when more dimensions are considered for a regression problem,
modeling the $X$-distribution might be too computationally expensive
or too inaccurate. In such a case, our correction by the 
$Y$-distribution 
is still feasible and improves the state of the art in regression evaluation.

\subsection{Evaluation on Robotic Data}
\label{s:rob}

\begin{figure*}[t!]
  \centering
  \includegraphics[width=\linewidth]{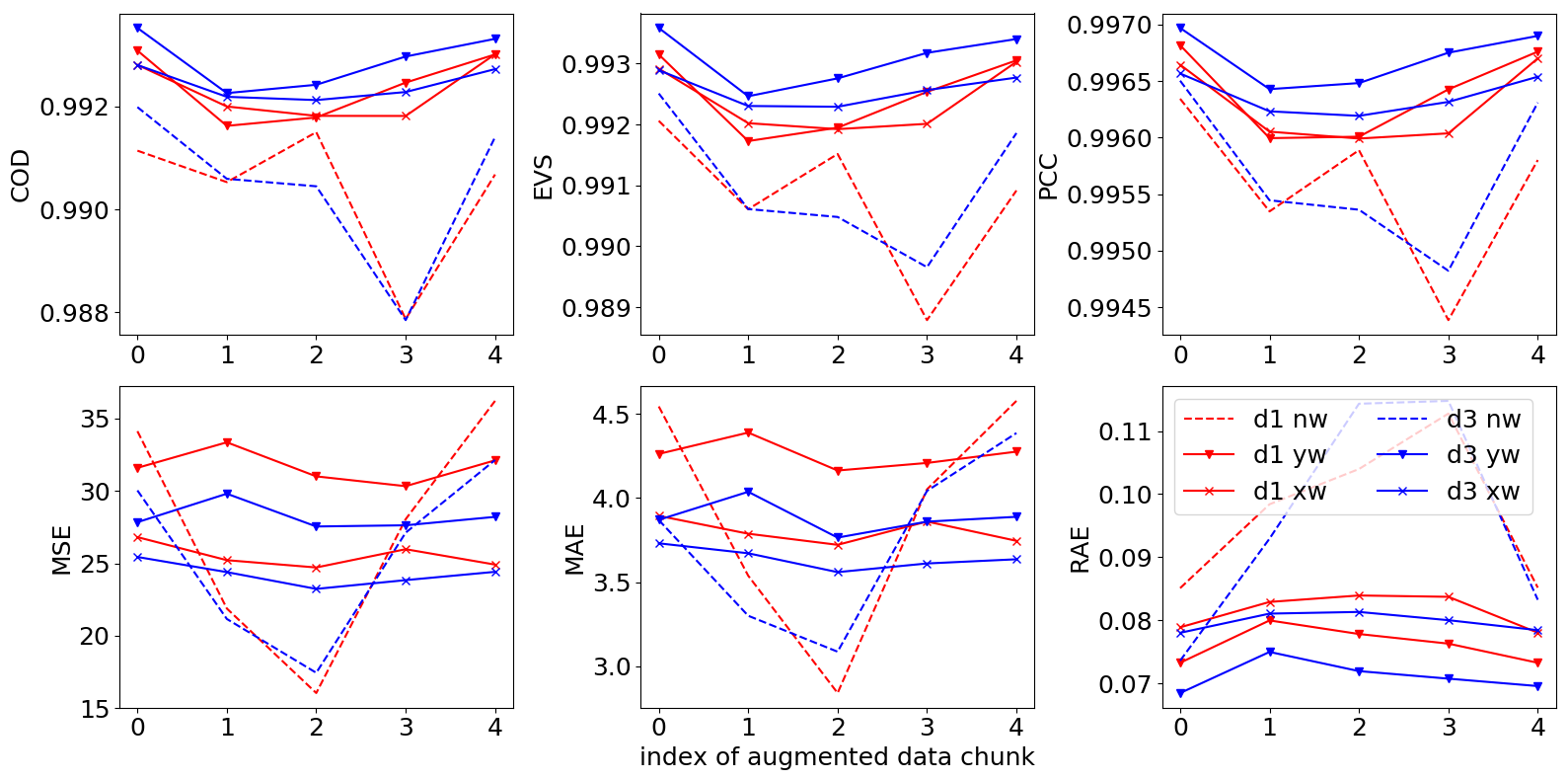}
  \caption{Comparison between classical metrics without weighting 
  (denoted by nw)
  and \emph{our} weight correction 
  using the X-distribution (\emph{xw})
  as well as the Y-distribution (\emph{yw})
  on dataset 1 (d1) and the reformulated dataset 3 (d3)
  for modeling the damping.
  The x-axis displays the index of the 
  overrepresented chunk~(see Section~\ref{s:rob}).
  A distribution invariant metric would be constant for all indices.
  }
  \label{f:robeval}
\end{figure*}

In this section, we introduce a real world example 
where we learn the damping term or resistive force
of a UUV in one degree of freedom (DOF). Following the definition of \cite{fossen02}, the damping term is
modeled as a function of the robot's velocity. For simplicity,
we only tackle the yaw DOF of the damping term,
which therefore reduces the equation to the following:
%
$d(\nu) = \tau - I_z\dot{\nu}$,
where $(\nu,\, \dot{\nu})$ are the robot's velocity and acceleration respectively, $\tau$ is the control effort due to actuation,
and $I_z$ the the robot's moment of inertia in the yaw DOF\footnote{
We used this equation to map dataset 3 to a damping estimation task
as for dataset 1.}.
The regression task can be summarized as fitting a function $f$
that maps $f(x_i) = y_i$, where $[x=\nu,\, y= d(\nu)] $.
Similar to Section~\ref{s:synth}, we use the support vector regression but with $C=10$ 
such that the function roughly fits the given data~\cite{Supplement}.
For testing, we generated different distributions by dividing
the samples ($x$) into $5$ equally sized chunks after sorting 
($x=[x_a,\ldots,x_e]$).
With every chunk we generated $5$ new augmented datasets
by using the original training data
and adding the data from the respective chunk $5$ times 
to obtain a distribution that
has a peak at the chunk's values 
(e.g., $x_0 = [x_a,\ldots,x_e, x_a, x_a, x_a, x_a, x_a]$)~\cite{Supplement}.
The $Y$-values were mapped correspondingly.
This resulted in $1800$ samples in total for dataset 1 and $2000$ for dataset 2.

The results are depicted in Fig.~\ref{f:robeval} for correction by
$X$- as well as by $Y$-distribution.
Note that for this example, the data is much more noisy and the 
distributions are difficult to fit.
Again, it can be clearly seen that there are very large variations 
of the classical uncorrected metrics depending on the distribution
indicated by the data chunk index.
Also, the weight corrected variants are much less sensitive to the
change in distribution
(but still not perfect).
Note that this time $X$-weighting and $Y$-weighting perform 
equally good/insensitive.
The minor peaks at the boundaries are due to very steep peaks
in the distributions that cannot be modeled well enough by the KDE.
This makes clear that it is important
to also take a look at the distributions and the respective KDE fit.
Here, maybe other kernels would have to be used or the histogram.


\section{Conclusion and Future Work}
\label{s:conc}

In this paper, we showed that non-uniform distributions of target variables 
in regression tasks have an effect on the established regression metrics
that cannot be ignored (up to $100\%$ difference).
To handle this imbalance, we propose several steps 
that should be followed to overcome this problem.
First, the distribution of actual regression values
should be visualized.
Second, a KDE, e.g., with a Gaussian kernel if it provides sufficient accuracy, 
should be fitted to the input samples.
Therefore, the efficient cross-validation scheme based on the 
integrated mean square error (``cv\_ls'') from the statsmodels library
could be used.
Third, the distribution fit is used as a (hyperparameter free) methodology 
to adapt existing evaluation metrics to imbalanced data
by using the inverse density values as weights in the
the definitions of established metrics.
%
The results on synthetic and
robotic data from UUVs 
show that our modified metrics compensate for the imbalance
almost perfect.
If a corrections by the $X$-distribution is not possible,
the $Y$-distribution can be used instead and still
provides much better results than the state of the art.

In future, we would like to look into other applications such as
location estimation and origami difficulty estimation.
We want to analyze more functions with more dimensions as input,
or output, and with sparse distributions.
Last but not least,
we want to incorporate our new evaluation concept 
into the loss function
of support vector regression or deep neural networks.





\section*{APPENDIX}
\label{s:app}

This section lists the detailed formulas for the new distribution invariant
regression metrics.

\begin{eqnarray}
  \mathit{MSE}_{\mathit{g}}(a,p) &=& \frac{
      \sum \limits_{i=1}^n g(a_i)^{-1}\cdot(p_i-a_i)^2}{
      \sum \limits_{i=1}^n g(a_i)^{-1}}\\
  \mathit{RMSE}_{\mathit{g}}(a,p)  &=&  \sqrt{\mathit{MSE}_{\mathit{g}}(a,p)}\\
  \mathit{MAE}_{\mathit{g}}(a,p)  &=&  \frac{
      \sum \limits_{i=1}^n g(a_i)^{-1}\cdot\left|p_i-a_i\right|}{
      \sum \limits_{i=1}^n g(a_i)^{-1}}\\
  \mathit{RSE}_{\mathit{g}}(a,p)  &=&  \frac{
      \sum\limits_{i=1}^n g(a_i)^{-1}\cdot(p_i-a_i)^2}{
      \sum\limits_{i=1}^n g(a_i)^{-1}\cdot(a_i-\bar{a})^2}\\
  \mathit{RRSE}_{\mathit{g}}(a,p)  &=&  \sqrt{\mathit{RSE}_{\mathit{g}}(a,p)}\\
  \mathit{RAE}_{\mathit{g}}(a,p)  &=&  \frac{
      \sum\limits_{i=1}^n g(a_i)^{-1}\cdot\left|p_i-a_i\right|}{
          \sum\limits_{i=1}^n g(a_i)^{-1}\cdot\left|a_i-\bar{a}\right|}\\
  \mathit{PCC}_{\mathit{g}}(a,p)  &=&  
  \frac{\sum\limits_{i=1}^n g(a_i)^{-1}\cdot(p_i-\bar{p})(a_i-\bar{a})
  }{\sqrt{
  {\sum\limits_{i=1}^n \frac{(p_i-\bar{p})^2}{g(a_i)}\cdot
  \sum\limits_{i=1}^n \frac{(a_i-\bar{a})^2}{g(a_i)}}
  }
  }\\
  \mathit{EVS}_{\mathit{g}}(a,p) &=& 1 - \frac{
      \sum\limits_{i=1}^n \frac{((p_i-a_i)-(\bar{p}-\bar{a}))^2}{g(a_i)}
  }{
  \sum\limits_{i=1}^n g(a_i)^{-1}\cdot(a_i-\bar{a})^2
  }
\end{eqnarray}

\bibliographystyle{ACM-Reference-Format}
\bibliography{library}

\end{document}